\documentclass{ifacconf}

\usepackage{graphicx}      
\usepackage{natbib}        

\usepackage{amsmath} 
\usepackage{amssymb}  
\usepackage{euscript}
\usepackage{xcolor}
\usepackage{svg}
\usepackage{bm}
\usepackage{optidef}
\usepackage{url}
\usepackage{algorithm}
\usepackage{algpseudocode}
\usepackage[shortlabels]{enumitem}
\usepackage{acronym}
\usepackage{multirow}
\usepackage{url}

\newtheorem{assumption}{Assumption}

\acrodef{PTP}{point-to-point}
\acrodef{OCP}{Optimal Control Problem}
\acrodef{ILC}{Iterative Learning Control}
\newcommand{\commentsymbol}{//}
\algrenewcommand\algorithmiccomment[1]{\hfill \commentsymbol{} #1}
\makeatletter
\newcommand{\LineComment}[2][\algorithmicindent]{\Statex \hspace{#1}\commentsymbol{} #2}
\DeclareMathSymbol{\smin}{\mathbin}{AMSa}{"39}
\begin{document}
\begin{frontmatter}

\title{Vibration Free Flexible Object Handling with a Robot Manipulator Using Learning Control} 

\author[First]{Daniele Ronzani$^1$} 
\author[First]{Shamil Mamedov$^1$}
\author[First]{Jan Swevers}

\address[First]{MECO Research Team, Dept. of Mechanical Engineering, \\KU Leuven, Belgium (e-mail: firstname.lastname@kuleuven.be)\\ and DMMS lab, Flanders Make, Leuven, Belgium.}

\begin{abstract}                
Many industries extensively use flexible materials. Effective approaches for handling flexible objects with a robot manipulator must address residual vibrations. Existing solutions rely on complex models, use additional instrumentation for sensing the vibrations, or do not exploit the repetitive nature of most industrial tasks. This paper develops an iterative learning control approach that jointly learns model parameters and residual dynamics using only the interoceptive sensors of the robot. The learned model is subsequently utilized to design optimal \ac{PTP} trajectories that accounts for residual vibration, nonlinear kinematics of the manipulator and joint limits. We experimentally show that the proposed approach reduces the residual vibrations by an order of magnitude compared with optimal vibration suppression using the analytical model and threefold compared with the available state-of-the-art method. These results demonstrate that effective handling of a flexible object does not require neither complex models nor additional instrumentation.
\end{abstract}


\thanks{These authors contributed equally.\newline$^\star~\, $This research was supported by the FWO-Vlaanderen through SBO project ELYSA for cobot applications (S001821N).}
\end{frontmatter}

\section{Introduction}

Innovative solutions in many industries require lighter, more durable, and often, consequently, flexible materials \citep{Saadat2002IndustrialApplications}. Applying standard solutions from rigid object manipulation to objects made from novel flexible materials lead to large vibrations. Existing feedback solutions require accurate sensing of the vibrations using an additional sensors and complex analytical or data-driven models. On the other hand, existing feedforward solutions increase the task execution time \citep{Singer1990ishaping}. Therefore, the industry can substantially benefit from new effective, yet simple solutions for flexible object handling.

In this paper we address the general problem of manipulating a flexible beam with a rigid robot arm \citep{Kapsalas2018ARXbeam}. 
We focus on solutions that do not use exteroceptive sensors for sensing vibrations of the beam -- such as external force-torque sensors at the end-effector or position tracking system -- only a joint torque estimator, available in the manipulator software, is used. Recently \cite{mamedov2022OBH} showed that using simple pendulum approximation of the beam and trajectory optimization, they can handle flexible objects better than existing methods. However, some residual vibration were still present.

Assuming that the beam handling is repetitive, this paper extends the work by \cite{mamedov2022OBH} and investigates whether vibrations can be further reduced by \ac{ILC}. A typical ILC algorithm uses the output error of the current task execution to update the input of the next run \citep{Bristow2006}. Robotic manipulators have been a common application for such learning techniques since its first mention \citep{Arimoto1984} to more recent advances \citep{Koc2019}. Generating a feasible input for robot manipulators with \ac{ILC} requires the algorithm to cope with nonlinear dynamics and hard joint constraints. \cite{Wang2018} used a filter-based \ac{ILC} with linearized model that demands a robust $\mathcal{H}_{\infty}$ design to account for such approximation.  \cite{Steinhauser2017} obtained feasible trajectories with an optimization-based ILC formulation where the the nonlinear dynamics and joint constraints were directly accounted for. In this paper we adopt a similar optimization-based strategy. Specifically, the problem at hand requires designing a \ac{PTP} trajectory for the manipulator which does not result in residual vibrations of the beam. Several \ac{ILC} techniques for optimizing  \ac{PTP} trajectories are available, e.g.  \cite{Freeman2011}, \cite{Son2013}, however they do not consider residual vibrations after motion. In contrast, \cite{VanDeWijdeven2008} proposed a vibration suppression \ac{ILC} that is, however, based on a predefined trajectory. Nonetheless, their method accounts for residual vibrations by formulating the problem with a separate control and prediction horizon similar to the proposed \ac{ILC}.


This paper proposes a vibration suppression \ac{ILC} for flexible object handling with a robot manipulator. The approach exploits the generic formulation from \cite{Volckaert2013} of an explicit learning and control steps, shown to be equivalent to a norm-optimal \ac{ILC}. The learning step consists of two estimation problems: the first, to learn a simple yet effective parametric model that approximates the flexible beam and considers the nonlinear kinematics of the robot manipulator; the second, to learn an equivalent output disturbance to account for the residual dynamics. Finally, in the control step, we formulate a vibration suppression \ac{OCP} for \ac{PTP} motions that exploits the learned dynamics and accounts for input and joints limits. Namely, we make the following contributions:
\begin{itemize}
    \item a measurement model for the external torque induced by a flexible object at the end-effector that accounts for the estimation error of the external torque provided by the manipulator software;
    \item a generalization of \ac{OCP} formulation from \citep{mamedov2022OBH} that leverages the learned residual dynamics and exploit a time-optimal-like formulation to induce zero residual vibration;
    \item experimental validation of the \ac{ILC} scheme.
\end{itemize}

This paper is organized as follows: Section \ref{sct:modeling} addresses the modeling of the robot arm, beam and external torque sensing. Section \ref{sct:ilc} discusses the proposed \ac{ILC} algorithm. Section \ref{sct:experiments} presents experimental results, followed by a discussion. Section \ref{sct:disc_conc} concludes the paper.



\section{Modeling}
\label{sct:modeling}
The vibration suppression \ac{OCP} \ac{PTP} motion controller requires a system model. Flexible objects are infinite dimensional systems; they are accurately modeled by partial differential equations (PDE) that are computationally demanding to solve and are seldom used in control and trajectory optimization. In robotics, for computationally tractable modeling of flexible objects, researchers make simplifying assumptions to convert PDE to ordinary differential equations \citep{Sakawa1985, Zhou2002NonlinearIsh}. The model parameters in the above-mentioned methods are obtained from CAD models because otherwise, in practice, it is difficult to estimate them. Data-driven methods approach modeling beam dynamics differently; they infer the model structure from data \citep{Kapsalas2018ARXbeam}. For modeling the beam we adapt the simple lumped modeling approach from \cite{mamedov2022OBH} and briefly describe it in this section for completeness.   


\subsection{Manipulator dynamics}
For a robot arm with $n_\mathrm{dof}$ degrees of freedom ($\mathrm{dof}$), let $\boldsymbol q \in \mathbb{R}^{n_\mathrm{dof}}$ be the vector of joint positions and assume that: \begin{assumption}\label{as:arm_double_int}
\label{as:arm_model}
The robot joint controller can accurately track the given joint reference trajectories.
\end{assumption}
Then, a double integrator model suffices to accurately describes the manipulators dynamics: 
\begin{align}  \label{eq:kin_model}
    \ddot{\boldsymbol q} = \boldsymbol u,
\end{align}
where $\ddot{\boldsymbol q} \in \mathbb{R}^{n_\mathrm{dof}}$ is the vector of joint accelerations, and $\boldsymbol u \in \mathbb{R}^{n_\mathrm{dof}}$ is the vector of inputs (reference joint accelerations).


\subsection{Beam dynamics on the end-effector}
\begin{figure}
    \centering

    \includegraphics[width=\linewidth]{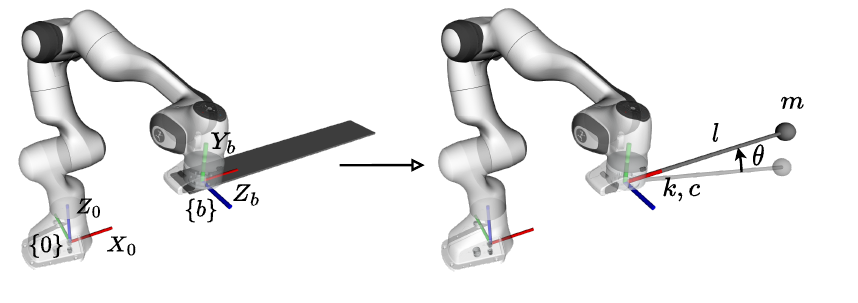}
    \caption{Approximation of a beam attached to the end-effector of a robot arm with a simple pendulum of length $l$ and a lumped mass $m$ connected to the end-effector through a passive revolute joint with stiffness $k$ and damping $c$.}
    \label{fig:modeling_assump}
\end{figure}
For modeling the beam dynamics manipulated by a robot arm, we make another critical and simplifying assumption: 
\begin{assumption}
\label{as:beam_model}
The beam can be approximated by a simple pendulum of mass $m$ and lenght $l$ connected to the end-effector of a robot arm through passive revolute joint with stiffness $k$ and damping $c$, as shown in Fig. \ref{fig:modeling_assump}. 
\end{assumption}
By making such assumption, we consider only the first natural frequency of a beam and only the lateral vibrations.

To derive the pendulum dynamics using the Lagrange formulation \cite[Ch. 7]{sciavicco2001book}, 
let $\bm p_{m}^0 \in \mathbb{R}^{3}$ denote the position of the pendulum mass $m$ in the robot's base frame
\begin{align} \label{eq:mass_pos}
    \bm p_{m}^0 (\bm q, \theta) = \bm p_{b}^{0}(\bm q) + l \bm R_{b}^{0} (\bm q) \bm R_z(\theta) {\bm i},
    %
\end{align}
where $[\bm p_{b}^{0}\ \bm R_{b}^{0}] = \mathrm{fk}(\bm q)$ are the position and the orientation of the origin of frame $\{b\}$, connected to the end-effector,  in base frame $\{0\}$, respectively and are obtained from the forward kinematics of the manipulator, $\bm R_z(\theta) \in \mathrm{SO(3)}$ is a rotation matrix around $Z_b$ axis, $\theta$ is the angular position of the pendulum and ${\bm i} = [1\ 0\ 0]^\top$ is a unit vector. From now on, we drop superscript $({}^0)$ and explicit dependence of variables on joint positions $\bm q$ and velocities $\dot{\bm q}$ for convenience. Using \eqref{eq:mass_pos} and its time derivative, it is possible to formulate the Lagrangian. 
%
Finally, applying the Lagrange equations leads the final expression for the pendulum dynamics
\begin{equation}\label{eq:pend_dynamics}
    \begin{split}
    \ddot \theta =& ~f_p(\bm q,\, \dot{\bm q},\, \theta,\, \dot \theta ) \\
    = & -\frac{1}{m\,l^2} ~\big(c \dot \theta+ k \theta\big) + \frac{1}{l}{\bm i}^\top \frac{d \bm R_z(\theta)}{d \theta}^\top \bm R_b^\top (\bm g - \ddot{\bm p}_{b}) \\
    & -{\bm i}^\top \frac{d \bm R_z(\theta)}{d \theta}^\top \bm R_b^\top \bm S(\dot{\bm \omega}_b) \bm R_b \bm R_z(\theta) {\bm i} \\
    & +{\bm i}^\top\frac{d \bm R_z(\theta)}{d \theta}^\top \bm R_{b}^\top \bm S(\bm \omega_b)^\top \bm S(\bm \omega_b) \bm R_{b} \bm R_z(\theta){\bm i}, 
    \end{split}
\end{equation}
where $\bm \omega_b \in \mathbb{R}^3$ and $\dot{\bm \omega}_b \in \mathbb{R}^3$ are the angular velocity and acceleration of frame $\{b\}$ with respect to $\{0\}$ expressed in $\{0\}$ respectively, $\bm S(\dot{\bm \omega}_b) := \dot{\bm R}_{b}  \bm R_{b}^\top  \in \mathbb{R}^{3\times 3}$ is a skew-symmetric matrix and $\bm g = [0\ 0\ -9.81]^\top\ \mathrm{m}/\mathrm{s}^2$  is the gravity acceleration vector. 


\subsection{External torque sensing}
\label{sct:modeling_tau_ext}
%
%
Any control strategy that attempts to improve the manipulation of the flexible beam requires measurements or estimates  of the beam motion in response to the control actions. Hence, in this subsection we develop an output model which complements the setup dynamics model from \cite{mamedov2022OBH}.


In absence of exteroceptive sensors, the beam dynamics can be inferred from the torque that its motion generates at frame $\{b\}$ along the $Z_b$ axis (see Fig. \ref{fig:modeling_assump}). Following the pendulum approximation of the beam (\ref{eq:pend_dynamics}), this reaction torque in frame $\{b\}$ along the $Z_b$ axis is written as:
\begin{align} \label{eq:beam_torque}
    \tau := \tau_{b, z}^b = - c \dot \theta - k \theta 
\end{align}
The software available in the robot manipulator drive system provides a filtered version of the external joint torque estimates $\bm \tau_{\mathrm{ext}}$ \citep{mamedov2020practical,petrea2021interactionForce} that is based on the dynamic model of the robot and torque measurement either at the joint or motor side. Therefore, a filtered version $\hat{\tau}_{b, z}$ in (\ref{eq:beam_torque}) is retrieved by using $\hat{\bm \tau}_{\mathrm{ext}}$ and the robots kinematics to compute the external wrench $\hat{\bm F}_b$ at the $\{b\}$:
\begin{align}
    \label{eq:torque_ext_to_wrench}
    \boldsymbol J_{b}^b(\boldsymbol q)^\top \boldsymbol \hat{\bm \tau}_{\mathrm{ext}} = \hat{\bm F}_b^b = [ \hat F_{b,x}^b\  \hat F_{b,y}^b\  \hat F_{b,z}^b\  \hat \tau_{b,x}^b\ \hat\tau_{b,y}^b\ \hat\tau_{b,z}^b]^\top
\end{align}
where $\boldsymbol J_{b}^b$ is the manipulator Jacobian in the $\{b\}$ frame.
As our variable of interest $\tau$ (\ref{eq:beam_torque}) can only be retrieved from its filtered version $\hat \tau :=\hat\tau^b_{b, z}$, we make the following output modeling assumption: 
\begin{assumption}
\label{as:meas_model}
The available output measurement is the external torque estimate in the frame $\{b\}$ along the $Z_b$ axis, filtered with a first-order low-pass filter: 
\begin{equation}
    \label{eq:output_model_filter}
\begin{split}
    \dot{\hat\tau} &= f_\tau(\hat \tau, \tau, \tau_{\mathrm{e}})= - a \hat \tau + a (\tau + \tau_{\mathrm{e}}) \\
     y &= \hat \tau
\end{split}
\end{equation} 
where $a$ is the inverse of the time constant of the filter and $\tau_e$ is a torque error given by assuming the following:
\end{assumption}
\begin{assumption}
\label{as:est_dyn}
The external torque estimator might not be correctly initialized but it converges exponentially.
\end{assumption}
\begin{align}
    \label{eq:output_model_error}
    \dot \tau_{\mathrm{e}} = -b \,\tau_{\mathrm{e}} \qquad \text{with} ~~ \tau_{\mathrm{e}}(0) = \tau_{\mathrm{e}, 0}
\end{align}
where $\tau_{e,0}$ is the unknown initial estimator error.
\subsection{Setup dynamics}
The setup model describes the dynamics later used by the learning algorithm to accomplish the task at hand. For this purpose the model is enhanced with a disturbance $d$ that affect the reaction torque \eqref{eq:beam_torque} as
\begin{align} \label{eq:beam_torque_d}
    \tau :=  - c \dot \theta - k \theta + d,
\end{align}
in order to capture the residual dynamics. Also, the dependency of the dynamics on a parameter $\bm p$ is made explicit, resulting in a model of the form $\dot{\bm x} = f(\bm x, \bm u, \bm p, d)$ and output map $y=h(\bm x, \bm u, \bm p, d)$. The setup model combines the manipulator dynamics \eqref{eq:kin_model}, the beam dynamics \eqref{eq:pend_dynamics}, the reaction torque filtering dynamics \eqref{eq:output_model_filter}-\eqref{eq:beam_torque_d}.
\begin{subequations}
    \label{eq:setup_model}
    \begin{align}
    \dot{\bm x} & = [\dot{\bm q}^{\top} ~~ \dot\theta ~~ \bm u^{\top} ~~ f_p(\cdot) ~~ f_\tau(\cdot) ~~ -b\, \tau_{\mathrm{e}}]^{\top} \label{eq:setup_model_dyn_f}\\
    y &=  ~ \hat \tau \label{eq:setup_model_dyn_h}
    \end{align}
\end{subequations}

where $\bm x = \big[\bm q^T ~~ \theta ~~ \dot{\bm q}^T ~~ \dot \theta ~~ \hat \tau ~~\tau_e \big]^T ~ \in \mathbb{R}^{n_x}$ is the state of the system with dimension $n_x = 2\,(n_{\mathrm{dof}}+1)+2$, $\bm p = [k~~ c~~ m~~ l~~ a~~ b~~ \tau_{\mathrm{e}, 0}]^\top$ is the vector of the parameters of the system and $\bm u$ is the control input as shown in \eqref{eq:kin_model}.
In the rest of this paper, we use discretized the setup dynamics $\bm x_{k+1} = \bm F(\bm x_{k}, \bm u_{k}, \bm p, \bm d)$ -- obtained from \eqref{eq:setup_model_dyn_f} using a $4$th-order Runge-Kutta integrator -- and the output map $y_{k} = H(\bm x_{k}, \bm u_{k}, \bm p, \bm d) := h(\cdot)$ obtained from \eqref{eq:setup_model_dyn_h} . 

 
\section{Iterative Learning Control}
\label{sct:ilc}
This section introduces the overall structure of the proposed \ac{ILC} algorithm for vibration free handling of a flexible object and subsequently details the two separate steps of the approach. We use the following notation: $(\cdot)^i$ denotes a particular iteration $i \in \mathbb{Z}_+ $  of the \ac{ILC}; $(\cdot)_k$ denotes a particular time sample $k \in \mathbb{Z}$ and $\bar{(\cdot)}$ indicates that the variable is pre-computed and/or given. 
\subsection{Algorithm/outline of the approach}
\begin{algorithm}
 \caption{Vibration free flexible object handling ILC}
 \begin{algorithmic}[1]
  \State\small\textbf{Require:}  $\bm p^0,\, d^0$ \Comment {prior parameters and disturbance}
  \State $\bm u^{1}\gets$ \small\verb|ocp(|$\bm p^0, \, d^0$\verb|)|
  \State $i \gets 1$
  \While{$i \leq i_{max}$}
  \State $\tilde{y}^i\gets$ \small\verb|system_response_measurement(|$\bm u^i$\verb|)|
  \LineComment[1\dimexpr\algorithmicindent]{Learning step}
  \State $p\bm ^i\gets$ \small\verb|parameter_estimation(|$\tilde{y}^i, \bm u^i,\bm p^{i\smin1}$\verb|)|\label{ALG_param_estimation}
  \State $d^i\gets$ \small\verb|disturbance_estimation(|$\tilde{y}^i, \bm u^i,\bm p^{i},d^{i\smin1}$\verb|)|\label{ALG_disturbance_estimation}
  \LineComment[1\dimexpr\algorithmicindent]{Control Step}
  \State $\bm u^{i+1}\gets$ \small\verb|ocp(|$\bm p^i, \, d^i, \, \bm u^i$\verb|)| \label{ALG_ocp}
  \State $i \gets i+1$
  \EndWhile
 \end{algorithmic}
 \label{ALG_ilc}
\end{algorithm}
Algorithm \ref{ALG_ilc} shows the general outline of proposed \ac{ILC}. It start with generating the first control input $\bm u^1$ is based on the given priors $\bm p_0$ and $d_0$. Next, the algorithm proceeds by iterating between: collecting the system response measurements $\tilde y^i$,$\bm u^i$, learning the parametric and residual dynamics $\bm p^i$, $d_i$ and computing the next control action $\bm u^{i+1}$ for vibration free handling.


\subsection{Learning step}
Traditionally, \ac{ILC} learns from the tracking error to update the next input. In the proposed approach, the learning is performed by explicitly correcting the model and learning the residual dynamics given the current experiment data $\bm u^i$, $\tilde{y}^i$.

In the first learning step, the model parameters $\bm p^i$ are obtained by solving the following nonlinear least-square estimation problem:
\begin{subequations}
\label{eq:estimation_problem_p}
\begin{align}
 \underset{\bm x, \theta_0,\, \bm p}{\text{min}} ~~ & \sum_{k=0}^{N-1}\big[\lVert\tilde{y}^i_k - y^i_k\rVert^2_2+ \underbrace{\lVert \bm p \rVert^2_{\bm V_1}}_{\mathrm{r}_{p,1}} + \underbrace{\lVert \bm p -\bm p^{i\smin1} \rVert^2_{\bm V_2}}_{\mathrm{r}_{p,2}}\big]\label{eq:est_objective}\\
 \text{s.t} \quad
    & \bm x_{k+1}= \bm F\big(\bm x_k, \bm u^i_k, \bm p\big), \quad k = 0,\dots,N-1, \label{eq:est_state_dyn}\\
    & y_k = H\big(\bm x_k, \bm u^i_k, \bm p\big),   \qquad k = 0,\dots,N-1,\label{eq:est_output}\\
    & f_{\mathrm{p,eq}}(\bm \bar{\bm q}_0, \theta_0, \bm p ) = 0, \label{eq:est_eq_state}\\
    & \bm x_0 = [\bar{\bm q}_0^\top\ \theta_{0}\ \bm 0^\top\ \hat \tau_0\ \tau_{e,0}]^\top,\label{eq:est_init} \\
    &\bm p \in \mathcal{P}\label{eq:est_p_constr}
\end{align}
\end{subequations}
where $\mathcal{P}$ is a feasible sets for the parameters, $f_{\mathrm{p,eq}}(\bm q, \theta, \bm p):= f_p(\bm q, \bm 0, \theta, 0, \bm p)$ and $\theta_0$ is the equilibrium position of the pendulum.
%
The main objective is to minimize the prediction error of the parametric model, i.e. (\ref{eq:est_state_dyn}) and (\ref{eq:est_output}) refer to the setup model (\ref{eq:setup_model}) where the disturbance $d$ is ignored. 

In the objective (\ref{eq:est_objective}) two regularization terms are added: $\mathrm{r}_{p,1}$, known as Tikhonov or Ridge regression \citep[Chp. 6.3.2]{Boyd2004convex},  improves the conditioning of the problem but introduces a bias; $\mathrm{r}_{p,2}$ regularizes the change in iteration domain to decrease the learning rate and hence to improve robustness against non-repetitive components, such as noise.


The second step in the learning procedure consists of capturing -- as an equivalent disturbance $d^i$ -- the residual dynamics that cannot be described by the parametric model. This is achieved by the following estimation problem where the model parameters are now set to the estimate $\bm p^i$ from the previous step:
\begin{subequations}
\label{eq:estimation_problem_d}
\begin{align}
 \underset{\bm x, \, \bm d}{\text{min}} ~~ & \sum_{k=0}^{N-1}\big[\lVert\tilde{y}^i_k - y^i_k\rVert^2_2
 + \underbrace{\lVert d_k \rVert^2_{w_1}}_{\mathrm{r}_{d,2}} +\\
 &\quad ~+ \underbrace{\lVert d_k- d^{i\smin 1}_k \rVert^2_{w_2}}_{\mathrm{r}_{d,2}}\big] + \sum_{k=0}^{N-2}\underbrace{ \big\lVert d_{k+1} - d_k \rVert^2_{w_3}}_{\mathrm{r}_{d,3}} \label{eq:est_d_objective}\\
 \text{s.t} \quad
    & \bm x_{k+1}= \bm F\big(\bm x_k, \bm u^i_k, \bm p^i, d_k\big), ~~ k = 0,\dots,N\smin1, \label{eq:est_d_state_dyn}\\
    & y_k = H\big(\bm x_k, \bm u^i_k, \bm p^i, d_k\big),   \qquad k = 0,\dots,N\smin1, \label{eq:est_d_output} \\
    & \bm x_0 = [\bar{\bm q}_0^\top\ \theta^i_{0}\ \bm 0^\top\ \hat\tau^i_{0}\ \tau^i_{e,0}]^\top. \label{eq:est_d_init}
\end{align}
\end{subequations}
Similar to (\ref{eq:estimation_problem_p}), regularization terms are added to the main objective that minimizes the prediction error. $\mathrm{r}_{d,1}$ penalizes the magnitude of the disturbance i.e. prevents $d_k$ from becoming too large. $\mathrm{r}_{d,2}$ increases robustness and regulates the learning rate. An additional regularization term $\mathrm{r}_{d,3}$ is added in(\ref{eq:est_d_objective}) to penalize the rate of change in time domain of the disturbance. This term imitates a low-pass filtering effect on the disturbance estimate and increases robustness w.r.t. measurements and process noise \citep[Ch. 6.3.2]{Boyd2004convex}.

\subsection{Control step}
\label{sec:ilc_control}
The vibration free flexible object handling task consists of a \ac{PTP} motion between two resting pose of the flexible beam connected to the end-effector. Such task is defined by the initial rest pose of the setup, determined by $\bar{\bm q}_{0}$ and $\bar{\theta}_{0}$; and final rest pose $\bar{\bm p}_{b, f}$, $\bar{\bm R}_{b, f}$, determined by $\bar{\bm q}_{f}$, with the corresponding equilibrium of the pendulum $\bar{\theta}_{f}$.
We compute the feedforward joint acceleration $\bm u_{i+1}$ by solving the \ac{OCP} that follows while using the current learned model information $\bm p_i$, $d_i$:
\begin{subequations}
\label{eq:ocp}
\begin{align}
 \underset{\bm x, \bm u}{\text{min}} ~~ & \phi_{c}(\bm x, \bm u, \bm u^i) +\phi_{p}(\theta, \dot \theta , \tau) \label{eq:ocp_objective}\\
 \text{s.t} \quad
    & \bm x_{k+1}= F\big(\bm x_k, \bm u^i_k, \bm p^i, d^i_k\big), ~~ k = 0,\dots,N_p\smin1, \label{eq:ocp_state_dyn}\\
    & \tau_{k} =  - k \, \theta_k -c\, \dot\theta_k + d^i_k \label{eq:ocp_tau}\\
    & \bm x_0 = [\bar{\bm q}_0^\top\ \bar{\theta}_{0}\ \bm 0^\top]^\top,  \bm u_0 = \bm0,\label{eq:ocp_init} \\
    & \bm p_{b}\left(\bm q_{N_c}\right) = \bar{\bm p}_{b, f},\ \dot{\bm q}_{N_c} = \bm 0,\\
    &\bm u_{k} = \bm 0, \qquad\qquad\qquad k=N_c\smin1,\dots, N_p\smin1, \label{eq:ocp_u_tf_zero}\\
    &\bm e_O\left(\bm R_{b}\left(\bm q_{N_c}\right), \bar{\bm R}_{b, f}\right) = \bm 0_{3\times 1}\\
    &\bm x \in \mathcal{X},\ \bm u \in \mathcal{U}, \ \dot{\bm u} \in \mathcal{J}.
\end{align}
\end{subequations}
where $\tau_k = \tau_{b,z,k}+d_k$ is the prediction of the pendulum reaction torque with the equivalent disturbance $d_k$ of the residual dynamics; 
$\bm e_O(\cdot) \in \mathbb{R}^3$ is a function for computing the orientation error between two frames \cite[Ch.3]{sciavicco2001book}; $\mathcal{X}$, $\mathcal{U}$, and $\mathcal{J}$ are feasible sets for states, controls, and rate of change of controls. 
In the problem formulation (\ref{eq:ocp}), we consider a control horizon of $N$ samples in which the motion is executed and for which a control horizon cost term in (\ref{eq:ocp_objective}) is designed to enforce a desirable motion of the robot manipulator:
\begin{equation}
\label{eq:ocp_objective_c}
\begin{split}
\phi_{c}(\cdot) = \sum_{k=0}^{N_c}\lVert \bm x_k - \bm x_0 \rVert^2_Q &+ \sum_{k=0}^{N_c-1}\lVert\bm u_k \rVert^2_{R_1}+\\
                 &+\sum_{k=0}^{N_c-2}\lVert \bm u_{k+1} - \bm u_{k} \rVert^2_{R_2},
\end{split}
\end{equation} 
where $\bm{Q} \in \mathbb{R}^{n_x \times n_x}, \bm R_1 \in \mathbb{R}^{n_{\mathrm{dof}} \times n_{\mathrm{dof}}}$, $\bm R_2 \in \mathbb{R}^{n_{\mathrm{dof}} \times n_{\mathrm{dof}}}$ are the weights for penalizing deviation of states from the initial state (to avoid excessive movements of the robot), inputs, prior input and jerks, respectively.
Additionally, we consider an extended prediction horizon from $N_c$ to $N_p$, in which the controls are set to zero (\ref{eq:ocp_u_tf_zero}), that is used to penalize the residual vibration occurring after the motion by means of the prediction horizon cost in (\ref{eq:ocp_objective}):
\begin{equation}
\label{eq:ocp_objective_p}
\begin{split}               
 \phi_{p}(\cdot)  = \sum_{k=N_c}^{N_p-1} \gamma^{k} \Big[\underbrace{\rho_1 \lVert \theta_{k} - \bar{\theta}_f\rVert_{1}}_{\mathrm{o}_{1}} &+ \underbrace{\rho_2 \lVert \dot{\theta}_{k}\rVert_{1}}_{\mathrm{o}_{2}} + \\
                &+ \underbrace{\rho_3\lVert \tau_{k} - \bar{\tau}_{f}\rVert_{1}}_{\mathrm{o}_{3}}\Big]
\end{split}
\end{equation}

The residual vibration are observed through $\theta_k$, $\dot\theta_k$ and $\tau_k$ and hence all three are considered in (\refeq{eq:ocp_objective_p}), each with their own weight $\rho_1$, $\rho_2$ and $\rho_3$ respectively.
The objective terms $\mathrm{o}_{1}$ and $\mathrm{o}_{2}$ penalize the prediction of the residual vibration by the parametric model. The aim is to keep $\theta_k$ close to the equilibrium position $\bar{\theta}_f$ and $\dot \theta_k$ equal to zero during the time horizon following the robot motion. Additionally, the term $\mathrm{o}_{1}$ penalizes the residual vibration as predicted by the torque $\tau_k$ (\ref{eq:ocp_tau}) which includes the residual dynamics given by $d^i_k$. To achieve vibration suppression $\tau_k$ should be equal to the equilibrium reaction torque given by $\bar{\tau}_f = -k\,\theta_f+\frac{1}{N_p-N}\sum^{N_p-1}_{k=N} d_k$.
Finally, note that all the terms in (\ref{eq:ocp_objective_p}) employ the sparsity promoting $l_1$-norm and are weighted by an exponentially increasing weight $\gamma>1$. This is done with the purpose of promoting zero residual vibration as early as possible after finishing the robot motion, that is as close as possible after reaching time instance $N_c$. A similar strategy is adopted in \citep{Verschueren2018} for a time-optimal model predictive control formulation.

\subsection{Numerical implementation}

In this work, we use CasADi \citep{Andersson2019casadi} to formulate the optimization problems ($\ref{eq:estimation_problem_p}$), ($\ref{eq:estimation_problem_d}$) and ($\ref{eq:ocp}$) as nonlinear programs (NLP) following the multiple-shooting method. 
The NLPs are solved using the nonlinear optimization solver IPOPT \citep{wachter2006ipopt} which implements an interior-point method. Moreover, we retrieve the computations of velocities and accelerations of the end-effector -- i.e, first- and second-order kinematics required in ($\ref{eq:estimation_problem_p}$), ($\ref{eq:estimation_problem_d}$) and ($\ref{eq:ocp}$)-- from the forward pass of the recursive Newton-Euler algorithm, which exploits the sparsity of the kinematic model unlike algorithmic differentiation. Such efficient functions for kinematics (and their derivatives) are generated by using Pinocchio \citep{carpentier2019pinocchio}. The code used in this work is publicly available on a GitHub repository\footnote{\url{https://github.com/danieleR3/beam_handling_ilc}}.

\section{Experiments}
\label{sct:experiments}
In this section, we describe the experimental setup, the task and the ILC settings. Then, we present the experimental validation of the proposed approach and compare it with an existent solution.

\begin{figure}[t]
    \centering
    \includegraphics[]{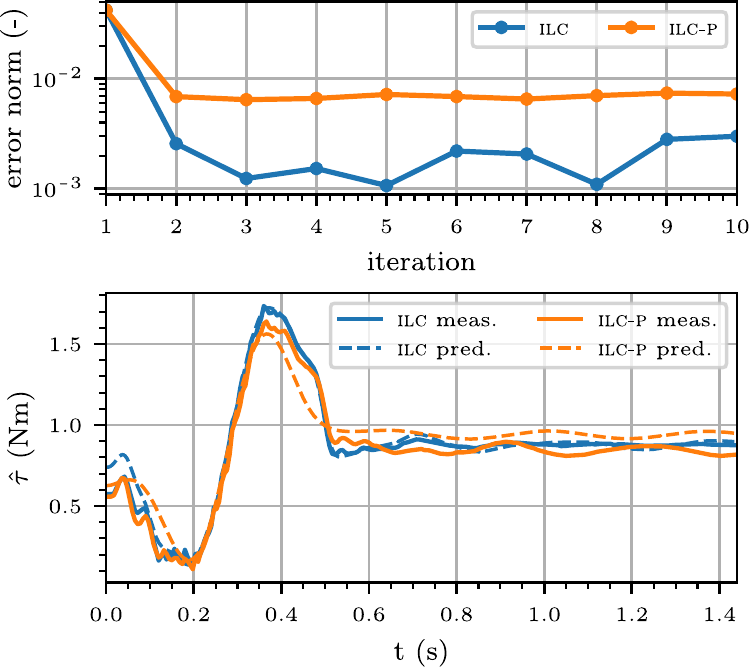}
    \caption{Top: error norm along the ILC iterations between the predicted, from \eqref{eq:ocp}, and the measured output. Bottom: comparison of the measured and prediction output for \textsc{ilc} and \textsc{ilc-p} at the $10$-th iteration.}
    \label{fig:T1_plot_learning}
\end{figure}

\subsection{Setup description}
The setup used to validate our approach consists of a 7-$\mathrm{dof}$ Franka Emika Panda manipulator and a flexible beam with dimensions $60\times6\times0.1 \ \mathrm{cm}$  rigidly attached to the arm's end-effector. 
The beam is made of stainless steel 316L with $\rho = 6.3\ \mathrm{g}/\mathrm{cm^3}$, $EI=1.267\ \mathrm{N}\cdot \mathrm{m}^2$.  
The actual inputs to the setup are reference joint velocities $\dot{\boldsymbol q}_r(t)$ retrieved by integrating the joint accelerations $\bm u_i$. The given outputs from the setup are joint positions $\boldsymbol q(t)$, velocities $\dot{\boldsymbol q}(t)$ and estimated filtered external torques $\hat{\boldsymbol \tau}_{\text{ext}}(t)$ at 1$\mathrm{kHz}$ as detailed in section \ref{sct:modeling_tau_ext}.


\subsection{Task definitions and ILC settings}

To demonstrate the functioning and the effectiveness of the proposed \ac{ILC} we consider the following beam handling task: starting from $\bm q_{0} = [ -\frac{\pi}{2}, -\frac{\pi}{6}, 0, -\frac{2\pi}{3}, 0, \frac{\pi}{2}, \frac{\pi}{4}]^\top$ move the end-effector by $[0.20\ 0\ -0.20]^\top \ \mathrm{m}$ relative to $\{0\}$ within $0.48\ \mathrm{s}$.
The \ac{ILC} algorithm is initialized with $\bm p^0$ obtained analytically from the beam material properties, as detailed in \citep{Sakawa1985}, and $d^0 =0$. The estimation problems \eqref{eq:estimation_problem_p} and \eqref{eq:estimation_problem_d} consider a horizon of $N=240$ samples with integration interval of $6\cdot10^{-3}\ \mathrm{s}$. Likewise, the control and prediction horizons in \eqref{eq:ocp} consist respectively of $N_c=48$, $N_p=144$ samples with integration interval of $10^{-2}\ \mathrm{s}$. 


The proposed ILC approach is compared with the $\textsc{baseline}$ approach, described in \citep{mamedov2022OBH}, that represent a special case of the \ac{OCP} (\ref{eq:ocp}) where only the parametric model is considered. The model parameters used in $\textsc{baseline}$ were determined by means of a data-driven method that rely on several ad hoc experiments.

To quantify the performance of the experiments we define as  metric the normalized integral of the absolute value of the zero mean residual vibrations (vibrations that persist after the end of the motion)
\begin{align}
    V = \frac{1}{N_r}\sum_{k=N}^{N + N_r} \left|\hat \tau - \bar{\hat{\tau}}\right|,
\end{align}
where $\bar{\hat{\tau}}$ is the average value of $\hat \tau$ and $N_r$ are samples of a sufficiently long time horizon such that it contains several of its periods in case of significant vibrations. In this paper, we consider a time window of $5\ \mathrm{s}$ in addition to the task motion time.

\begin{figure}[!t]
    \centering
    \includegraphics[]{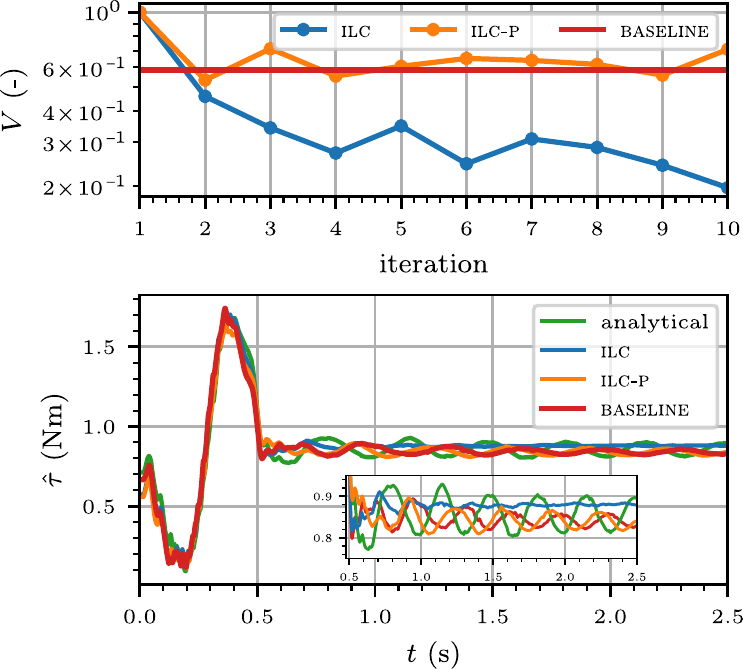}
    \caption{Top: comparison of the vibration performance metric along the ILC iteration. Bottom: comparison of the residual vibrations induced in the measurements $\hat \tau$ for the first and last iteration of \textsc{ilc} and \textsc{ilc-p} and for the \textsc{baseline}.}
    \label{fig:T1_plot_performance}
\end{figure}
\subsection{Validation}

The proposed \ac{ILC} algorithm combines a parametric model and a disturbance that represents the residual dynamics. To understand its functioning, we perform Algorithm \ref{ALG_ilc} (\textsc{ilc}) and compare it to the case where the parameter estimation does not include the residual dynamics (\textsc{ilc-p}), i.e., $d^i =  0$.
Figure \ref{fig:T1_plot_learning} shows that by combining the parametric and the disturbance models, the \textsc{ilc} more accurately predicts the output with respect to \textsc{ilc-p}, especially the residual vibrations. This result motivates the need to learn the residual dynamics and leverage it via the extended prediction horizon cost \eqref{eq:ocp_objective_p}. 
Figure \ref{fig:T1_plot_performance} compares the performance of both  \textsc{ilc}, \textsc{ilc-p} and the \textsc{baseline}. The top figures shows the evolution of the residual vibrations as a function of the ILC iterations. \textsc{ilc} achieves nearly zero residual after a short time interval, especially compared to the first experiment that exploits the analytical model. Despite of that, \textsc{ilc-p} still achieve a considerable reduction of the vibration w.r.t the initial experiment and obtains a comparable vibration suppression to \textsc{baseline}. Note that \textsc{ilc} and \textsc{ilc-p} learn the model parameter by exploiting the execution of the task, while \textsc{baseline} requires ad-hoc experiments prior to the task. A visual demonstration of the experiments can be found at \url{https://youtu.be/c8vi91NDlkg}.

\section{Conclusion}
\label{sct:disc_conc}
This paper proposes an ILC algorithm for vibration free flexible object handling with a robot manipulator. Assuming that the beam handling is repetitive, this paper extends the work by \cite{mamedov2022OBH}. We present a measurement model for the external torque induced by the flexible object that accounts for the estimation error introduced by the manipulator software. The model enables learning of a parametric model and residual dynamics without relying on any exteroceptive sensors. Unlike other ILC approaches, the proposed algorithm introduces a PTP optimal control strategy that accounts for residual vibration, nonlinear kinematics and physical limits of the manipulator. The approach is experimentally validated  and shows a threefold improvement compared with the available state-of-the-art method. This result is mainly due to estimating and exploiting the residual dynamics. This work can provide a solution for learning \ac{PTP} motion primitives useful for executing more challenging and industrially relevant handling tasks. 

\bibliography{references}   
\end{document}